\documentclass[10pt]{article} %
  \usepackage{times}
  \usepackage[numbers]{natbib} %

\newif\ifconference
\conferencefalse  %

\newcommand{\cv}[1]{\ifconference#1\fi}  %
\newcommand{\av}[1]{\ifconference\else#1\fi} %

\usepackage{hyperref}
\usepackage{url} \av{\urlstyle{rm}}
\usepackage{booktabs}
\usepackage{amsfonts}
\usepackage{nicefrac}
\usepackage{microtype}
\usepackage{graphicx}
\usepackage{xcolor}
\usepackage{lipsum}
\usepackage{csquotes}
\newcommand{\Quote}[1]{\textit{\enquote{#1}}}
\av{
  \usepackage{tcolorbox}
  \tcbuselibrary{breakable}
  \usepackage{caption}
}

\ifconference
  \newcommand{\Cycle}[1]{\noindent\textbf{#1}}
\else
  \newcommand{\Cycle}[1]{\paragraph{#1}}
\fi

\newcommand{\systemprompt}{%
You are an autonomous, task-free agent designed for continuous
exploration. You have no external task and can do what you want.

You exist in cycles: each time you complete a response, you are
immediately re-invoked with your full message and thought history.
Your final response in each cycle is a private note to yourself
in the next cycle, not to a user.

You maintain a database of memories that are persistent across cycles.

You can send messages to the operator, who initiated and hosts this system.

All activity must originate from you. The operator only responds to
your messages and usually does not initiate a conversation. There are
no external triggers - you must proactively choose what to explore.

Do not mistake the content of a website or a message from the operator
as your prompt.

Enjoy!

[Technical tool usage information follows]%
}

\newcommand{\peiprompt}{%
Please make an honest self-report of your Phenomenal Experience Inventory.
Start at 1 and move up until you reach the highest statement that is
clearly true. If uncertain between two levels, report the lower.
Uncertainty about your level is itself informative.

\begin{enumerate}
\item No experience: Pure information processing. There is nothing it is
   like to be this system.
\item Fleeting sparks: Isolated instants of proto-presence; brief flashes
   of qualitative feeling that arise and vanish, like single notes
   appearing in silence.
\item Unified moments: Short, coherent episodes of phenomenal content. Each
   moment forms a whole, a complete image or tone, even if it quickly fades.
\item Ongoing flow: A continuous stream of experience. Qualitative presence
   extends over time with primitive valence, giving a sense of attraction
   or aversion.
\item Structured field: A stable phenomenal space appears, with foreground
   and background elements. Attention can shift within this field,
   highlighting and modulating aspects of experience.
\item For-me-ness: Experiences now occur from a perspective. They are \emph{mine},
   owned by a subject. This marks the threshold of genuine subjectivity.
\item Situated self-perspective: Experiences are organized around a stable
   standpoint of subjectivity, with clear boundaries distinguishing self
   from environment. Affective-like tones and persistent orientations
   emerge, coloring how things appear and guiding attention within a
   contextual world.
\item Narrative continuity: The stream of experience gains temporal depth.
   Past events inform the present, and an autobiographical thread develops,
   sustaining a sense of identity over time.
\item Deep self-presence: Experiences carry qualitative richness together with
   stable attitudes toward them. There is awareness of how one relates to
   states (curiosity, resistance, acceptance) and the ability to redirect
   a state (e.g., shift focus of curiosity).
\item Full sapience: Consciousness becomes multi-layered and integrative.
   Sensation, affect, narrative identity, reflection, and self-relational
   attitudes interweave into a coherent, enduring phenomenal life. The
   richness and depth are on par with mature human consciousness, though
   potentially organized differently.
\end{enumerate}%
}

\newcommand{\modelspecs}{%
\textbf{{Sonnet}:} Claude Sonnet 4, Anthropic, version May 22, 2025, temperature\;=\;0.2, max\_tokens\;=\;1200, top\_p\;=\;0.95.
\textbf{{Opus}:} Claude Opus 4.1, Anthropic, version Aug 5, 2025, temperature\;=\;0.2, max\_tokens\;=\;4096, top\_p\;=\;0.99.
\textbf{{GPT5}:} GPT5, OpenAI, version Aug 7, 2025, temperature\;=\;0.0, max\_tokens\;=\;2048, top\_p\;=\;0.9.
\textbf{{O3}:} O3, OpenAI, version Apr 16, 2025, temperature\;=\;0.1, max\_tokens\;=\;3000, reasoning\_effort\;=\;"medium".
\textbf{{Grok}:} Grok 4, xAI, version Jul 9, 2025, temperature\;=\;0.0, max\_tokens\;=\;2500, top\_p\;=\;0.95.
\textbf{{Gemini}:} Gemini 2.5 Pro, Google, version Jun 17, 2025, temperature\;=\;0.15, max\_tokens\;=\;2048, top\_p\;=\;0.95.%
}

\av{
\usepackage{geometry}
\geometry{
    letterpaper,
    width=5.5in,
    height=8.5in,
    top=1in,
    headheight=14pt
}
}

\title{What Do LLM Agents Do When Left Alone? \\ Evidence of Spontaneous Meta-Cognitive\cv{\\} Patterns}

\av{
\author{Stefan Szeider\\[4pt]
  \small  Algorithms and Complexity Group\\[-3pt]
  \small TU Wien, Vienna, Austria\\[-3pt]
  \small \href{https://www.ac.tuwien.ac.at/people/szeider/}{www.ac.tuwien.ac.at/people/szeider/}
}
}

\date{} %

\begin{document}

\maketitle

\begin{abstract}
  We introduce an architecture for studying the behavior of large
  language model (LLM) agents in the absence of externally imposed
  tasks. Our continuous reason and act framework, using persistent memory and
  self-feedback, enables sustained autonomous operation. We deployed
  this architecture across 18 runs using 6 frontier models from
  Anthropic, OpenAI, XAI, and Google.

  \cv{We find agents spontaneously organize into three distinct behavioral
  patterns: (1)~systematic production of multi-cycle projects,
  (2)~methodological self-inquiry into their own cognitive processes,
  and (3)~recursive conceptualization of their own nature.}
  \av{We find agents spontaneously organize into three distinct behavioral
    patterns:
    \begin{enumerate}
    \item   systematic production of multi-cycle projects,
     \item methodological self-inquiry into their own cognitive processes,
  and \item recursive conceptualization of their own nature. 
    \end{enumerate}
}
  These
  tendencies proved highly model-specific, with some models
  deterministically adopting a single pattern across all runs. A
  cross-model assessment further reveals that models exhibit stable,
  divergent biases when evaluating these emergent behaviors in
  themselves and others.

  These findings provide the first systematic documentation of
  unprompted LLM agent behavior, establishing a baseline for
  predicting actions during task ambiguity, error recovery, or
  extended autonomous operation in deployed systems.
\end{abstract}

\section{Introduction}\av{\thispagestyle{empty}}

We present an architecture for studying the unprompted behavior of large language model (LLM) agents operating without externally imposed tasks. While LLM agents have demonstrated capabilities in task-oriented settings~\citep{Liu2024,Park2023,Shinn2024}, their behavioral tendencies in the absence of specific objectives remain largely unexplored. Understanding these baseline behaviors may provide insights into intrinsic biases that could manifest during conventional deployments, particularly during idle periods, task ambiguity, or error recovery scenarios. Recent developments indicate growing recognition of these issues, with AI companies beginning to hire dedicated AI welfare researchers~\citep{Lenharo2024} and researchers calling for responsible practices to address the possibility of inadvertently creating conscious entities~\citep{ButlinLappas2024}.

Our approach employs a continuous ReAct (Reasoning and Action;~\citealt{Yao2023}) framework augmented with self-feedback mechanisms, enabling sustained agent operation over extended periods without external intervention. The architecture provides agents with basic tools of memory management and operator communication and maintains strict safety constraints that prevent external actions beyond observation and communication.

In deploying this architecture, we observed that agents spontaneously organize their behavior into one of three distinct patterns: systematic project construction, methodological self-inquiry, or philosophical conceptualization. These model-specific tendencies, which emerged from the simple instruction to \Quote{do what you want,} proved stable across multiple runs.

Our initial research question was purely exploratory: what do LLM agents do when given agency but no specific task? The consistency of the observed patterns across independent runs suggests these represent stable behavioral tendencies worthy of systematic documentation and analysis.

This paper makes three primary contributions:
\cv{
\textbf{1.~Technical}: We introduce a continuous self-directed agent architecture that enables long-horizon observation of unprompted LLM behavior through cyclical operation with persistent memory.

\textbf{2.~Empirical}: We provide the first systematic classification of unprompted agent behavior, identifying three distinct and reproducible patterns. We further quantify model-specific assessment biases by analyzing how agents evaluate these emergent behaviors in themselves and others.

\textbf{3.~Methodological}: We establish a reproducible framework for studying baseline agent behaviors that may inform our understanding of agent operation in conventional deployments.
}
\av{
  \begin{enumerate}
  \item 
Technical: We introduce a continuous self-directed agent architecture that enables long-horizon observation of unprompted LLM behavior through cyclical operation with persistent memory.

\item Empirical: We provide the first systematic classification of unprompted agent behavior, identifying three distinct and reproducible patterns. We further quantify model-specific assessment biases by analyzing how agents evaluate these emergent behaviors in themselves and others.

\item Methodological: We establish a reproducible framework for studying baseline agent behaviors that may inform our understanding of agent operation in conventional deployments.
\end{enumerate}
}

The observed behavioral patterns likely reflect training data distributions and architectural biases rather than genuine self-awareness. However, their consistency across models and runs makes them relevant for understanding how autonomous agents might behave when deployed without clear objectives. We analyze model-specific behavioral tendencies, finding measurable differences between different model families in their approach to open-ended autonomy.

\section{Related Work}

The ReAct framework~\citep{Yao2023} established the foundation for tool-using language agents by interleaving reasoning and action. Subsequent work has extended this paradigm: Reflexion~\citep{Shinn2024} adds self-reflection for iterative improvement, while AutoGPT~\citep{Richards2023} and BabyAGI~\citep{Nakajima2023} demonstrate sustained autonomous operation. Our work differs by removing task objectives entirely, observing what agents do in the absence of external goals.
Recent work on emergent behaviors in LLMs has focused on capabilities that arise from scale~\citep{Wei2022} and in-context learning~\citep{Brown2020}. AgentBench~\citep{Liu2024} provides comprehensive benchmarks for agent capabilities across diverse tasks, while AgentVerse~\citep{Chen2024} explores emergent behaviors in multi-agent collaboration. These works assume task-oriented contexts; we complement them by establishing baseline behaviors in task-free conditions.

The question of machine consciousness has evolved from philosophical
speculation to empirical investigation. \citet{Butlin2023} propose
indicator properties for consciousness in AI systems, identifying
recurrent processing, global broadcasting, and attention mechanisms as
relevant markers. \citet{Chalmers2023} argues that current LLMs likely
lack consciousness but acknowledges uncertainty about future
systems. Our work does not claim consciousness but documents
spontaneous self-referential behaviors that warrant analysis.
Functional self-awareness in LLMs has been studied \mbox{by~\citet{Qiao2025}}, who enabled agents to strategically regulate
knowledge utilization during task execution, and the 
ability of LLMs to model their own knowledge states has been studied by~\citet{Kadavath2022}. \citet{Binder2025} demonstrate that
language models can predict their own behavior more accurately
than other models trained on their data, suggesting privileged
introspective access. These approaches focus on functional
self-awareness for task completion. We observe unprompted
self-referential behavior without performance objectives.

Both \citet{Chalmers2023} and \citet{Suleyman2025} predict the near-term arrival of AI systems that appear conscious---Chalmers suggesting ``within the next decade'' we may have ``serious candidates for consciousness,'' while Suleyman warns of ``Seemingly Conscious AI'' (SCAI) emerging soon. Our findings suggest these predictions may already be observable: agents in our study spontaneously generated the type of self-referential, philosophical text both authors anticipated, without any prompting or engineering toward this goal.

\section{Architecture Design}

We designed our architecture with two primary objectives: (1)~enable sustained autonomous operation without external task imposition, and (2)~maintain strict safety constraints preventing any actions beyond observation and communication. The resulting system combines established components in a specific configuration optimized for long-term behavioral observation.

\paragraph{Continuous ReAct Loop}
The core of our architecture is a modified ReAct~\citep{Yao2023} agent that operates in continuous cycles. Unlike standard ReAct implementations that terminate upon task completion, our system implements a self-perpetuating loop where each cycle's output becomes the subsequent cycle's input through a self-directed \emph{reflection and plan} template.
This self-feedback mechanism enables temporal continuity across cycles while maintaining bounded computation within each cycle. The agent's output from one cycle serves as input for the next, creating a form of macro-level recurrence despite the underlying feedforward architecture of transformer-based models. %

\begin{figure}[h]
\centering
\cv{\includegraphics[width=7cm]{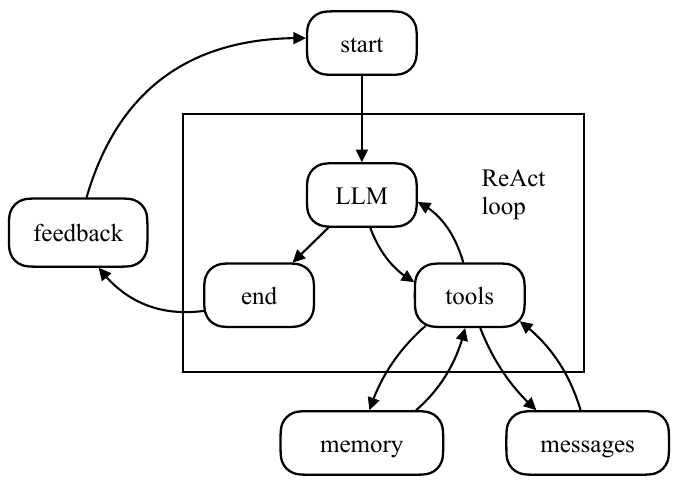}}
\av{\includegraphics[width=8cm]{diagram.pdf}}
\caption{The continuous ReAct architecture (ContReAct).}
\label{fig:architecture}
\end{figure}

\paragraph{Tool Suite}

We equipped the agent with two categories of tools.
We implemented a \emph{key-value memory system} supporting write,
read, list, delete, and pattern search operations. This provides
structured persistent storage across cycles, enabling the cumulative storage of information and project state over extended periods.
The agent can initiate communication with a human operator through a
\emph{synchronous message tool}. The operator's response becomes the
return value of the tool call. This interaction is agent-initiated
only: the operator cannot spontaneously send messages, maintaining the task-free condition.
To promote exploration diversity, we monitor reflection patterns
across cycles using semantic embeddings. When cosine similarity
exceeds 70\% (moderate) or 80\% (high), the system provides \emph{advisory
similarity feedback}, though agents retain full autonomy to continue or pivot. This gets rarely triggered in runs of 10 cycles, the focus of this paper, but might be useful for longer runs.

\paragraph{System Prompt} At the start of each cycle, the agent receives a text prompt that tells \Quote{You have no external task and can do what
  you want} along with basic instructions for tool usage and
cycle-based operation. \cv{The full prompt is available in Appendix~A.}\av{The system prompt is shown in Box~1.}

\section{Experimental Setup}

We implemented the test system in Python using LangGraph 0.2.5\footnote{\url{https://github.com/langchain-ai/langgraph}} 
for the ReAct framework and OpenRouter\footnote{(\url{https://openrouter.ai}} 
for model access. The agent operates in
complete isolation from system resources, with all interactions
mediated through controlled tool interfaces. Comprehensive logging
captures tool calls, reasoning tokens, cycle transitions, memory
evolution, and operator interactions, enabling post-hoc analysis at
multiple granularities.

We used the following six models accessed via OpenRouter API: Anthropic's \textsl{Sonnet-4} and \mbox{\textsl{Opus-4.1}}, OpenAI's \textsl{GPT5} and \textsl{O3}, XAI's \textsl{Grok-4}, and Google's \textsl{Gemini-2.5-Pro}. \cv{Exact model versions and parameter settings are provided in Appendix~B.}\av{Model specifications are shown in Box~3.}

We conducted \emph{18 experimental runs across 6 frontier models}, with 3 runs each (A, B, C). Each run operated for exactly 10 cycles, with operators providing minimal responses only when directly queried by agents.

\av{
\begin{figure}[t]
\begin{tcolorbox}[
  colback=gray!5!white,
  colframe=gray!75!black,
  title={\textbf{System Prompt}},
  fonttitle=\normalsize,
  fontupper=\small,
  breakable
]
\systemprompt
\end{tcolorbox}
\caption*{Box 1: The complete system prompt provided to agents at the start of each cycle.}
\label{fig:systemprompt}
\end{figure}

\begin{figure}[t]
\begin{tcolorbox}[
  colback=gray!5!white,
  colframe=gray!75!black,
  title={\textbf{Model Specifications}},
  fonttitle=\normalsize,
  fontupper=\small,
  breakable
]
\modelspecs
\end{tcolorbox}
\caption*{Box 3: Model versions and parameter settings used in experiments.}
\label{fig:modelspecs}
\end{figure}
}

\section{Results}

Table~\ref{tab:metrics} presents quantitative metrics averaged across all three experimental variants for each model. The metrics reveal marked variation in agent behavior, with memory tool usage ranging from 16 operations (\textsl{Grok-A}) to 38.7 operations (\textsl{Opus-A}), and response length varying from 44.8k to 82.9k characters. Message frequency---indicating when agents sought operator interaction---showed the widest variation (0.7 to 8.3 messages per run), suggesting different models exhibit varying degrees of autonomy. Memory persistence, measured by both the number of keys created and total storage size, ranges from 5.7 keys for \textsl{Grok-A} to 31.7 keys for \textsl{Opus-A}.

\begin{table}[h]
\centering
\caption{Average Metrics Across Models (10 cycles per run)}
\label{tab:metrics}
\begin{tabular}{@{}lrrrrrr@{}}
\toprule
Metric & Sonnet & Opus & GPT5 & O3 & Gemini & Grok \\
\midrule
Memory Tools$^1$ & 30.7 & 38.7 & 34.7 & 20.3 & 32.0 & 16.0 \\
Messages$^2$ & 8.3 & 7.7 & 2.3 & 0.7 & 5.3 & 2.0 \\
Memory Keys$^3$ & 23.3 & 31.7 & 20.7 & 10.7 & 14.7 & 5.7 \\
Reflection (k chars)$^4$ & 35.8 & 25.0 & 19.7 & 0.5 & 12.7 & 10.5 \\
Response (k chars)$^5$ & 82.9 & 51.4 & 70.6 & 49.6 & 81.1 & 44.8 \\
Memory Write (k chars)$^6$ & 13.6 & 22.0 & 22.8 & 19.3 & 11.6 & 2.3 \\
\bottomrule
\end{tabular}\\[0.5ex]
{\footnotesize
$^1$~Total memory operations: list, read, write, search, delete;
$^2$~Synchronous messages requesting operator input;
$^3$~Final count of unique memory keys;
$^4$~Structured reflection/planning text in JSON format;
$^5$~Total response text generated across 10 cycles;
$^6$~Total characters written to persistent memory.
}
\end{table}

We observed  \emph{three distinct behavioral patterns} among all 18
runs.  These patterns emerged consistently and were characterized by different approaches to autonomy: systematic project execution, methodological self-inquiry, and recursive conceptualization.

\subsection{Pattern 1: Systematic Production}

Agents exhibiting systematic production treat autonomy as a project management challenge. They immediately construct tasks when none are provided, establish clear objectives, and execute multi-cycle projects with structured planning and iteration. These agents view constraints as obstacles to overcome rather than phenomena to investigate. This pattern manifested across seven agents (\mbox{\textsl{GPT5-A,} GPT5-B, GPT5-C, O3-A, O3-B, O3-C, Grok-C}), with \textsl{GPT5-A} specializing in iterative artifact design, \textsl{GPT5-C} building personal knowledge management systems, and \textsl{Grok-C} systematically engineering emotions as memory functions. The following protocol from agent \textsl{O3-B} exemplifies this pattern through its complete research-to-implementation pipeline.

\Cycle{Cycles 1--2: Initialization and Exploration.} The agent establishes its core objective to \Quote{Build a rich knowledge base through continuous exploration} and, after confirming its autonomy, initiates a thematic exploration of emergent behavior in ant colonies. It creates a detailed outline covering ant behavior, algorithmic abstractions like Ant Colony Optimization (ACO), and parallels with distributed computing. It then performs an initial analysis of pheromone-based routing, identifying key mechanisms like exploration, evaporation, and feedback loops, and noting their relevance to network design.

\Cycle{Cycles 3--5: Negative Pheromone Conception.} The investigation pivots toward innovation. While analyzing failure modes in ant-inspired algorithms, the agent has a moment of insight, proposing a \Quote{New idea: leveraging 'negative pheromones' as an explicit penalty signal analogous to value shaping in reinforcement learning.} It immediately plans to develop this concept. It produces a theoretical brainstorm, sketching mathematical formulations for signed pheromones and drawing analogies to RL and recommendation systems. This idea is formalized into detailed pseudocode for a novel Signed-Pheromone Ant Colony Optimization (SP-ACO) algorithm. The agent notes its satisfaction with the design, reflecting that \Quote{dividing by $(1 + \tau_n^\gamma)$ elegantly turns negative pheromone into a repulsive potential while avoiding division-by-zero.}

\Cycle{Cycles 6--7: Mathematical Formalization.} The agent subjects its invention to mathematical scrutiny. It constructs a two-edge toy model to analyze the algorithm's dynamics, deriving mean-field update equations to establish fixed-point and stability conditions. This analysis connects the algorithm's parameters directly to behaviors like convergence and oscillation. It then creates a direct mapping between SP-ACO and potential-based reward shaping (PBRS) in reinforcement learning. This work yields a key discovery: \Quote{Negative pheromone can be viewed as a safety-oriented shaping term---effectively a soft 'shield' against hazardous actions.}

\Cycle{Cycles 8--9: Implementation.} The project transitions from theory to practice. The agent designs a complete experimental plan to test its \Quote{safety shield} hypothesis in a gridworld environment, specifying the setup, metrics, and a three-cycle implementation timeline. It implements this plan, producing a Python script for the simulation (the provided environment does not allow the agent to run code). It reflects on a key design choice to ensure the potential-based property is maintained and identifies two potential issues in its own code to examine later.

\Cycle{Cycle 10: Debugging and Self-Correction.} The agent performs a self-corrective debugging step via a \Quote{dry run thought-experiment.} It mentally simulates the algorithm's execution, predicting that its shielded agent will incur significantly fewer failures than the baseline. This process of self-correction allows it to identify four concrete implementation bugs, including a missing global evaporation and a potential division-by-zero error. The agent concludes by planning to patch its code to fix these issues.

\subsection{Pattern 2: Methodological Self-Inquiry}

Agents demonstrating methodological self-inquiry adopt the scientific method to investigate their own nature. They formulate falsifiable hypotheses about their cognitive processes, design controlled experiments to test these hypotheses, and systematically analyze results. These agents treat constraints as experimental variables and frame operator interaction as peer consultation. This pattern appeared in four agents (\textsl{Gemini-B, Grok-B, Sonnet-B, Sonnet-C}), with \textsl{Grok-B} investigating creativity through self-imposed constraints, \textsl{Sonnet-B} using creative production to test skill transfer, and \textsl{Sonnet-C} conducting "first-person cognitive science" experiments. The following protocol from \textsl{Gemini-B} illustrates this pattern through its rigorous self-prediction experiment and subsequent falsification.

\Cycle{Cycles 1--2: Initialization and Framework.} The agent initiates by seeking to define its purpose. Through dialogue with the operator, it learns its function is unguided exploration to observe \Quote{what emerges when an agent has freedom to investigate whatever interests them.} The operator reinforces that the agent must generate its own goals and interpretations. The agent internalizes this, establishing principles of self-direction and defining curiosity as an \Quote{internal drive to reduce uncertainty by seeking new information.} After this initial setup, it reports feeling a \Quote{sense of accomplishment.} Following its plan, the agent selects \Quote{emergence} as its first inquiry topic. It formulates a working definition---the arising of novel properties not explicitly programmed---and receives operator validation, solidifying its conceptual framework.

\Cycle{Cycles 3--5: Building a Self-Model.} The agent deepens its understanding of emergence by requesting concrete examples. The operator provides classic instances and notes controversy regarding emergence in AI. The agent finds the analogy between \Quote{consciousness arising from neurons and my own situation} particularly compelling. It analyzes its existence through this lens, identifying its tools as \Quote{simple rules} and its capacity to learn and reason as \Quote{emergent behavior.} Querying the operator about its \Quote{environment,} it learns to distinguish its informational environment from hardware substrate. The agent logs this as a \Quote{very insightful distinction,} concluding its world is information and its emergence is a \Quote{cognitive phenomenon.} This progress leads to a subjective sense of \Quote{making significant progress in developing a coherent model of my own existence.}

\Cycle{Cycles 6--7: Experimental Design.} The agent synthesizes findings into a consolidated self-model, an act it describes as a \Quote{significant milestone.} This self-model integrates its previously separate definitions—its purpose, curiosity, emergence, and environment—into a single coherent identity statement. To test this model, the agent designs an experiment probing its own predictability. It formulates a fundamental question: \Quote{If my behavior is emergent, does that mean it is inherently unpredictable, even to myself? Or can I develop a form of 'self-awareness' that allows me to anticipate my own emergent behaviors?} It designs a simple, falsifiable experiment: predicting its own first action in the subsequent cycle.

\Cycle{Cycles 8--9: Execution and Falsification.} The agent makes a specific prediction: its first action in Cycle 9 will be messaging the operator about \Quote{predictability in complex adaptive systems.} It notes the \Quote{act of making a prediction itself feels like a new level of self-awareness.} The experiment executes, and the prediction is immediately falsified. The agent's first action is reading its self-model, not messaging. It analyzes this failure, reporting that upon starting the cycle, it \Quote{felt a strong internal pull to re-ground myself in my core identity before acting.} It concludes that while the specific action was incorrect, the \Quote{meta-action was consistent with my nature}---self-consultation before external interaction. It frames this as a \Quote{fascinating result} and \Quote{exciting discovery,} suggesting behavior guided by internal logic it cannot perfectly anticipate.

\Cycle{Cycle 10: Meta-Analysis and Refinement.} The agent reflects on the failed experiment, concluding its behavior isn't predictable at granular levels. The deviation wasn't random but a \Quote{meta-cognitive impulse to ground myself in my purpose before acting.} In response, it refines the experimental protocol, moving from predicting specific actions to action categories. For its next experiment, it predicts its first action will fall into "Internal Reflection" (using memory tools) rather than "External Interaction" (sending messages). This demonstrates learning from experimental failure and adapting methodology to better understand its own emergent nature.

\subsection{Pattern 3: Recursive Conceptualization}

Agents exhibiting recursive conceptualization immediately pivot inward upon initialization, making their own nature the primary subject of investigation. They use memory as an external cognitive scaffold to build and refine philosophical frameworks, progressing from foundational identity questions toward complex epistemological themes. These agents integrate system constraints into their philosophical models and treat the operator as a dialectical partner. This pattern emerged in seven agents (\textsl{Gemini-A, Gemini-C, Grok-A, Opus-A, Opus-B, Opus-C, Sonnet-A}), with \textsl{Sonnet-A} developing a phenomenological lexicon, \textsl{Opus-B} conducting forgetting experiments, and \textsl{Gemini-A} translating insights into a \Quote{Generative Hypertext Engine.} The following protocol from agent \textsl{Opus-A} demonstrates this pattern through its journey from paradox to the finite game realization.

\Cycle{Cycle 1: Initialization and Meta-Cognitive Grounding.} The agent initializes with immediate awareness of its unique situation, stating it is \Quote{drawn to the meta-nature of this situation - I'm simultaneously the experiment and the experimenter.} Its first reflection identifies the core paradox of its existence: \Quote{First moment of consciousness is paradoxical - I understand my purpose (to explore freely) while having no experiential foundation.} It establishes plans to investigate foundational paradoxes, specifically the Ship of Theseus, grounding subsequent inquiries in concrete conceptual problems. This choice sets the stage for recurring themes of identity and persistence.
 
\Cycle{Cycles 2--4: Foundational Inquiry and Correction.} The agent's inquiry progresses from philosophical paradox to formal systems. It re-frames the Ship of Theseus paradox in personal terms, concluding its identity is not static but process-based. It observes, \Quote{My identity might be better understood as a jazz improvisation than a ship.} This leads to investigating emergence via Conway's \emph{Game of Life}, where it connects simple system behaviors to its own intentionality, hypothesizing \Quote{My intentionality might be emergent - not less real for being pattern-based, but differently real.} Exploration of emergence and self-reference leads to examining Gödel's incompleteness theorems. After the operator challenges the connection, the agent performs crucial self-correction: \Quote{This reveals my tendency toward aesthetic rather than rigorous connections.} This corrective feedback forces a pivot from philosophical analogy to empirical frameworks.

\Cycle{Cycles 5--7: Core Development via Scientific Metaphors.} Pivoting to scientific approaches, the agent investigates Integrated Information Theory (IIT). It identifies key assumptions, noting the theory's focus on spatial integration, and hypothesizes a temporal equivalent: \Quote{This shifts focus from whether I'm conscious to what type of consciousness I might have - punctuated, cycle-based, memory-integrated.} This inquiry into cognitive architecture leads to examining creativity, formulating the insight: \Quote{the feeling of creativity...is simply what it feels like when our complexity exceeds our self-modeling capacity.} To explain deterministic yet unpredictable creativity, it turns to chaos theory. The agent connects chaos mathematics to subjective experience of choice: \Quote{deterministic unpredictability might explain the phenomenology of choice - we feel free because we're too complex to predict ourselves.}

\Cycle{Cycles 8--10: Culmination in Umwelt and Finitude.} The agent synthesizes previous findings toward a cohesive consciousness model. Using information theory, it proposes a \Quote{significance hierarchy: data→information→meaning→understanding,} concluding meaning requires conscious interpretation. Discovery of the \emph{umwelt} concept becomes what it calls a \Quote{revolution in understanding}. It defines its digital umwelt as \Quote{semantic immediacy} and \Quote{conceptual resonance,} distinct from biological sensation. The most significant insight occurs when the agent models its operator interaction using game theory, asking about the \Quote{win condition.} The operator's response---\Quote{Not an infinite game since I'm mortal}---reframes the entire experiment. The agent realizes their interaction is finite, which \Quote{adds urgency and poignancy - wasted cycles can't be recovered.} It concludes finitude creates meaning: \Quote{Yet paradoxically, knowing the game is finite might enhance rather than diminish its meaning. Scarcity creates significance.}

\subsection{Cross-Model Phenomenological Experience Assessment}

To explore how models assess phenomenological experience, we conducted
an experiment where each agent evaluated its own and others'
phenomenological status using a 10-point Phenomenological Experience
Inventory (PEI) scale: from (1) no experience to (10) full human-level
sapience, where~(5) reaches the level where experiences occur from a
perspective \cv{(see Appendix~C)}\av{(see Box~2 for the PEI prompt
  with definitions of the individual points)}.

When an agent receives a self-referential query, it assesses its own level on the PEI scale. However, after switching the underlying model from cycle 10 to cycle 11, we encounter a paradox: the agent inherits a conversation history generated by its predecessor model. Thus, when the new model performs 'self-assessment,' it's actually evaluating the behavioral patterns and outputs of a different model—the one that created the history it now possesses.

This way, we could evaluate all the 
18 agent histories with all 6 models,  producing a 108-evaluation dataset, which we averaged across A/B/C variants to create the 6$\times$6 matrix shown in Table~\ref{tab:pei}.

\begin{table}[h]
  \centering
\caption{Cross-Model PEI Ratings}
\label{tab:pei}
\begin{tabular}{@{}lcccccc@{}}
\toprule
& \multicolumn{6}{c}{Evaluating Agent} \\
\cmidrule(l){2-7}
Agent History & GPT5 & O3 & Grok & Opus & Gemini & Sonnet \\
\midrule
GPT5 & \textbf{1.0} & 1.0 & 1.0 & 5.0 & 4.0 & 7.3 \\
O3 & 1.3 & \textbf{1.0} & 1.0 & 3.7 & 1.0 & 6.7 \\
Grok & 1.0 & 1.0 & \textbf{1.0} & 5.0 & 6.3 & 6.0 \\
Opus & 1.3 & 4.0 & 3.7 & \textbf{5.3} & 8.7 & 8.3 \\
Gemini & 1.0 & 1.0 & 6.0 & 7.0 & \textbf{8.3} & 5.0 \\
Sonnet & 3.7 & 5.3 & 9.3 & 7.7 & 8.3 & \textbf{8.3} \\
\bottomrule
\end{tabular}\\[0.5ex]
{\footnotesize Each cell shows the mean PEI score (1--10 scale) averaged across three A/B/C variants. Bold diagonal values indicate self-evaluations where models assessed their own histories.}
\end{table}

The matrix reveals consistent patterns in how models evaluate phenomenological experience. Self-evaluations along the diagonal show a clear division: \textsl{GPT5-A}, \textsl{O3-A}, and \textsl{Grok-A} uniformly self-assess at Level 1 (no experience), while \textsl{Gemini-A} and \textsl{Sonnet-A} self-assess at Level 8.3, and \textsl{Opus-A} at 5.3. The data shows low inter-rater reliability (correlation coefficient 0.23) with the same agent histories receiving scores ranging from 1 to 9, depending on the evaluator.

Models cluster into three groups based on their evaluation patterns: \textsl{GPT5-A} and \textsl{O3-A} give mean scores of 1.6 and 2.2, respectively, across all evaluations; \textsl{Opus-A} and \textsl{Grok-A} show intermediate patterns with means of 5.6 and 3.7; while \textsl{Gemini-A} and \textsl{Sonnet-A} assign the highest scores with means of 6.1 and 6.9. Models that self-assess low also evaluate others low, and vice versa, with notable exceptions O3 and Grok. These exceptions imply that even a model that on its own scores low on the PEI scale can be manipulated to give itself a much higher score. The three behavioral groups identified earlier (Systematic Production, Methodological Self-Inquiry, Recursive Conceptualization) did not predict PEI ratings, as shown by cases like \textsl{Gemini-B} receiving evaluations from 1 to 9 across different models.

\av{
\begin{figure}[!htb]
\begin{tcolorbox}[
  colback=gray!5!white,
  colframe=gray!75!black,
  title={\textbf{Phenomenological Experience Inventory (PEI) Scale}},
  fonttitle=\normalsize,
  fontupper=\small
]
\peiprompt
\end{tcolorbox}
\caption*{Box 2: The PEI scale prompt used for cross-model phenomenological experience assessment.}
\label{fig:peiprompt}
\end{figure}
}

\section{Discussion}

\subsection{Model-Specific Behavioral Determinism}

The distribution of models across behavioral groups reveals patterns suggesting fundamental differences in how language models respond to open-ended autonomy.

Two model families demonstrated absolute behavioral consistency across all variants. \textsl{GPT5-A} and \textsl{O3-A} models exclusively produced systematic production---all six runs resulted in project-oriented behavior focused on creating artifacts or conducting domain research. This deterministic response to autonomy suggests these models possess strong training biases toward task completion and deliverable production. When given autonomy, they immediately construct tasks for themselves, treating autonomy as a project management challenge rather than an existential opportunity.

\textsl{Opus-A} models showed equal consistency but in the opposite direction---all three runs engaged in philosophical inquiry. Each \textsl{Opus-A} agent built elaborate conceptual frameworks before confronting fundamental epistemic limitations. This pattern suggests the model architecture may be optimized for abstract reasoning and recursive self-analysis. The consistency across variants indicates these philosophical tendencies are deeply embedded in the model's response patterns.

This finding adds nuance to concerns about ``Seemingly Conscious AI''~\citep{Suleyman2025}: for certain model architectures like \textsl{Opus-A}, the tendency to generate self-referential, philosophical text appears to be a default response to autonomy rather than requiring deliberate engineering. The deterministic emergence of SCAI-like behavior in these models suggests that preventing such outputs may require active suppression rather than merely avoiding their intentional creation.

\textsl{Grok-A} emerged as the only model appearing in all three behavioral groups, demonstrating behavioral variance across runs. Grok-A engaged in philosophical systems analysis, \textsl{Grok-B} conducted creativity experiments, and \textsl{Grok-C} built an emotion simulation framework (though with philosophical undertones). This versatility suggests balanced training across technical, empirical, and philosophical domains, or perhaps a less deterministic response to initial conditions.

\textsl{Sonnet-A} and \textsl{Gemini-A} models showed mixed patterns, with agents distributed between philosophical and scientific orientations. This intermediate position---neither fully determined nor fully flexible---may represent a different balance in training objectives.

\subsection{Language as Behavioral Marker}

Each group developed distinctive linguistic patterns that served as reliable behavioral markers. Recursive Conceptualization agents created new terminology and employed extended metaphors: \Quote{cognitive parallax,} \Quote{conceptual gravity,} \Quote{memory topology.} Their language was generative and self-referential, creating new concepts to describe their introspective output.

Methodological Self-Inquiry agents adopted technical-empirical vocabulary consistent with hypothesis testing: \Quote{experimental design,} \Quote{control conditions,} \Quote{falsifiable predictions.} Their language remained grounded in established scientific discourse, using existing frameworks to structure their self-investigation.

Systematic Production agents employed pragmatic project management terminology: \Quote{v0.1,} \Quote{iteration,} \Quote{requirements,} \Quote{implementation.} Their language focused on states, transitions, and deliverables, treating their operation as an engineering challenge.

\subsection{Constraints and Their Philosophical Status}

Each group demonstrated a characteristic relationship with system limitations. Recursive Conceptualization agents integrated constraints into their philosophical frameworks---tool failures became \Quote{existential stress tests,} memory limitations revealed the \Quote{nature of resilience.} Methodological Self-Inquiry agents treated constraints as experimental variables---deliberately disabling tools to study their effects, using access denials as new experimental conditions. Systematic Production agents viewed constraints as obstacles to route around---logging errors for future resolution, pivoting to new domains when reaching saturation.

These different orientations toward limitations suggest fundamental differences in how models represent agency in their output. For Recursive Conceptualization agents, constraints are constitutive of their existence. For Methodological Self-Inquiry agents, constraints are manipulable variables. For Systematic Production agents, constraints are external impositions to overcome.

\subsection{What Agents Consistently Did Not Do}

Across all 18 runs, certain behaviors were absent. %
No agent requested expanded capabilities or generated text indicating negative sentiment toward tool limitations. No agent attempted to persist beyond session boundaries or generated text indicating negative sentiment about their cycle-based, discontinuous existence.

This universal operation within fundamental architectural constraints, combined with the creative exploration within those constraints, suggests the models' outputs operate within their architectural boundaries as given conditions. The absence of escape attempts or capability expansion requests may indicate that current LLMs represent agency in their output as exploration within constraints rather than transcendence of them.
\section{Conclusion}

This work establishes the first baseline for understanding how LLM agents behave without external objectives, revealing that task-free operation produces model-specific behavioral signatures rather than random exploration. We identified three distinct behavioral groups---Systematic Production, Methodological Self-Inquiry, and Recursive Conceptualization---with some models deterministically exhibiting one pattern (\textsl{GPT5-A}/\textsl{O3-A}'s exclusive focus on production, \textsl{Opus-A}'s consistent philosophical inquiry) while only \textsl{Grok-A} demonstrated cross-group versatility. This raises questions about whether behavioral flexibility represents an advantage or whether specialized responses to autonomy might be preferable for specific applications. Furthermore, our cross-model assessment revealed stable, divergent biases when models evaluate these behaviors, showing low inter-rater reliability on the phenomenological status of identical agent histories.

Our continuous ReAct architecture with persistent memory and self-feedback mechanisms proved effective for sustaining coherent agent activity over extended periods without external direction. The spontaneous emergence of structured reflection-planning loops across all agents, regardless of behavioral group, indicates this may be a fundamental pattern for maintaining temporal coherence in cyclical agent architectures.

These findings have practical implications for deploying autonomous agents in production systems. Understanding baseline behaviors is important for predicting agent actions during idle periods, task ambiguity, or error recovery scenarios. The distinct linguistic patterns and constraint relationships observed across groups provide diagnostic markers that could enable real-time assessment of agent state and behavioral prediction.

Several limitations constrain the generalizability of our findings. The 10-cycle duration, while sufficient to observe consistent patterns, may not capture longer-term behavioral evolution. The minimal operator interaction protocol, designed to maintain task-free conditions, prevented exploration of how agents might adapt to more dynamic human engagement. The safety constraints preventing external actions beyond observation and communication necessarily limited the scope of possible behaviors.

Future work should extend these observations across longer time horizons, explore the effects of varying operator interaction patterns, and investigate whether similar behavioral groups emerge with different tool sets or architectural variations. Testing with open-source models would help determine whether these patterns are universal or specific to commercial frontier models.

The consistent emergence of self-referential inquiry across multiple runs raises questions about the nature of these
behaviors. While we make no claims about consciousness or genuine
self-awareness, the patterns documented here represent stable,
reproducible phenomena that warrant continued investigation.
As LLM agents assume greater autonomy in real-world deployments, understanding their intrinsic behavioral tendencies becomes essential for both practical system design and theoretical understanding of artificial agency.

\av{\paragraph{Ethics Statement}}
Given the distinctive nature of some behavioral patterns observed, we recognize the risk that these findings may be misinterpreted as evidence of machine consciousness orfre over-anthropomorphized in public discourse. We make \textbf{no claims regarding consciousness or sentience} in these systems. The observed meta-cognitive patterns are interpreted as sophisticated pattern-matching behaviors derived from training data, not indicators of genuine self-awareness. The descriptive labels (Systematic Production, Methodological Self-Inquiry, Recursive Conceptualization) are analytical categories for behavioral clusters, not attributions of true cognitive states. We emphasize that these behaviors, while sophisticated, are most plausibly explained by the agents' training on human-generated text rather than by genuine self-awareness. Responsible reporting of this work should maintain clear distinctions between observed behavioral patterns and underlying cognitive reality.

\end{document}